\pdfoutput=1
\documentclass[sigconf, screen, nonacm]{acmart}
\usepackage{booktabs}
\usepackage{multirow}
\AtBeginDocument{%
  }
\renewcommand\footnotetextcopyrightpermission[1]{}

\begin{document}

\title{TRAP: Tail-aware Ranking Attack for World-Model Planning}

\author{Siyuan Duan}
\affiliation{%
  \institution{Department of Computer Science and Technology, Soochow University}
  \city{Suzhou}
  \country{China}
}

\author{Ke Zhang}
\authornote{Corresponding author.}
\email{kzhang19@suda.edu.cn}
\affiliation{%
  \institution{Department of Computer Science and Technology, Soochow University}
  \city{Suzhou}
  \country{China}
}

\author{Xizhao Luo}
\authornote{Corresponding author.}
\email{xzluo@suda.edu.cn}
\affiliation{%
  \institution{Department of Computer Science and Technology, Soochow University}
  \city{Suzhou}
  \country{China}
}

\renewcommand{\shortauthors}{Duan et al.}

\begin{abstract}
World models enable long-horizon planning by internally generating and evaluating imagined trajectories, making them a promising foundation for generalist agents. However, this imagination-driven decision process also introduces new security risks. Existing backdoor attacks typically aim to manipulate local features, one-step predictions, or instantaneous policy outputs. While such objectives may suffice for weaker reactive models, they are often ineffective against world models, where the learned dynamics prior and planning process can absorb or wash out the effects of shallow perturbations. More importantly, we find that world models exhibit a distinct backdoor vulnerability rooted in the long-tailed ranking structure of imagined trajectories, where disrupting the ordering of a few decision-critical trajectories can systematically hijack planning.

To exploit this vulnerability, we propose TRAP, a backdoor attack framework for world models that targets imagined trajectory ranking. TRAP combines a \textbf{tail-aware ranking loss} to focus optimization on decision-critical trajectories with \textbf{dual gating mechanisms} that stabilize optimization and regulate when and where the attack penalty is applied. Under trigger conditions, TRAP alters the relative ranking of imagined trajectories to redirect planning outcomes, while largely maintaining the normal ranking structure on clean inputs. Experiments on DreamerV3 and TD-MPC2 across diverse tasks show that TRAP consistently induces sustained behavioral deviations and significant performance degradation, highlighting the need for dedicated security evaluation of world-model-based agents.
\end{abstract}






\maketitle

\section{Introduction}
With the rapid development of world models~\cite{ha2018world,hafner2019dream,hafner2023mastering,hansen2023td,bruce2024genie,brooks2024video}, learning to model future evolution has become a powerful paradigm for intelligent systems and generalist agents. By generating \textbf{imagined trajectories} for \textbf{multi-step planning}, planning-based world models offer strong capabilities in long-horizon reasoning and generalization~\cite{hafner2019dream,hafner2023mastering,hansen2023td,chen2022transdreamer,zhang2023storm}. However, their reliance on evaluating large sets of \textbf{candidate futures} also introduces new security vulnerabilities, especially in high-stakes applications such as robotics and autonomous driving. This raises a critical question: can these imagined futures be systematically manipulated while preserving normal behavior on clean inputs?

Backdoor attacks are a \textbf{stealthy and persistent threat}: a compromised model \textbf{behaves normally on clean inputs}, yet can be manipulated when a \textbf{specific trigger} is present~\cite{gu2019badnets,li2022backdoor,kiourti2020trojdrl,cui2024badrl}. For world-model-driven agents, such attacks can be particularly consequential, since once the \textbf{planning process is compromised}, the resulting behavioral deviation may \textbf{persist over long horizons}. Studying backdoor attacks against world models therefore concerns not only model security, but also the reliability and safety of model-based decision systems. However, unlike conventional perception models, world models rely on \textbf{longer decision chains} and \textbf{structured future prediction}~\cite{hafner2019dream,hafner2023mastering,hansen2023td}, where the effect of local perturbations is mediated by learned dynamics and multi-step planning. This naturally raises the question of whether existing backdoor attacks can be directly applied to world-model-driven agents.

We find that they become \textbf{substantially less effective}---and that this failure is not incidental but structural. Existing attacks typically target one-step predictions, latent representations, or policy outputs~\cite{gu2019badnets,kiourti2020trojdrl,cui2024badrl,zhou2025badvla,ma2025unidoor}, whereas world models make decisions by generating and \emph{ranking} many candidate trajectories over long horizons~\cite{hafner2019dream,hafner2023mastering,hansen2023td,chen2022transdreamer,zhang2023storm}. This mismatch gives rise to two specific failure modes. First, because decisions depend on evaluating multiple imagined rollouts, local trigger-induced deviations must propagate through multi-step dynamics, reward accumulation, and value estimation before they can influence behavior. Consequently, such local anomalies are highly susceptible to \textbf{attenuation or even cancellation} over time, rendering conventional one-step attack objectives largely ineffective in the world-model setting. Second, existing attacks do not directly target a central decision mechanism in world-model planning: the \textbf{relative ordering} of candidate trajectories. Even when triggers produce noticeable shifts in latent states or action outputs, they often fail to reliably alter this ordering, because the planner's final action selection depends on the comparative ranking among all candidates rather than the absolute score of any single trajectory. This makes \textbf{sustained and reliable attacks} difficult under existing paradigms. These limitations highlight the need for a trajectory-aware attack formulation tailored to world-model planning.

\begin{figure}[t]
    \centering
    \includegraphics[width=\columnwidth]{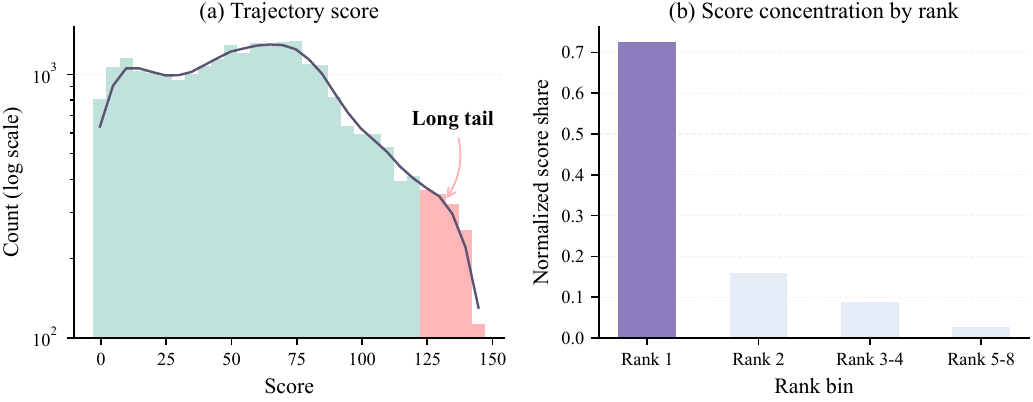}
    \caption{Motivation of TRAP.
(a) Under clean conditions, imagined trajectory scores exhibit a strongly skewed distribution, where only a small fraction lies in the high-score tail.
(b) After sorting trajectories by score, the normalized score share concentrates on the top-ranked trajectories, showing that only a few imagined futures dominate planning.}
    \label{fig:intro_longtail}
\end{figure}

Our key insight is that decision-making in world-model-driven reinforcement learning critically depends on the \textbf{relative ranking of trajectories}, rather than absolute prediction errors. During planning, the agent generates many candidate trajectories and selects actions according to their cumulative returns or value estimates~\cite{hafner2019dream,hafner2023mastering,hansen2022temporal,hansen2023td,chen2022transdreamer,zhang2023storm}. As illustrated in Fig.~\ref{fig:intro_longtail}, this planning process naturally induces a \textbf{long-tailed distribution} over trajectory scores, in which only a small number of high-value trajectories dominate decision-making. Consequently, even small but systematic shifts in the ranking of these trajectories can propagate through the planner and lead to stable behavioral deviations. Motivated by this observation, we propose TRAP, a tail-aware attack framework for world-model planning. TRAP operates in a deployment-time, trigger-conditioned setting, without relying on training-time poisoning or parameter tampering. Rather than perturbing one-step predictions, it directly targets the \textbf{relative ordering} of candidate trajectories. The key idea is to attack the planner where it is most sensitive: we first use a tail-aware ranking objective to identify and optimize against the small set of \textbf{decision-critical tail trajectories} that dominate planning, and then introduce two explicit gating losses to regulate when attack penalties should be enforced during optimization. In this way, the tail-aware objective determines \emph{which} imagined futures to attack, while the gating terms determine \emph{when} such attacks should be applied. This design enables TRAP to reliably manipulate planning outcomes while maintaining normal behavior on clean inputs.

In summary, our main contributions are as follows:

\noindent \textbf{(1)} We propose \textbf{TRAP}, the first trigger-conditioned, inference-time backdoor attack framework for world models that manipulates long-horizon decision-making via visual triggers, without requiring parameter tampering or training-time poisoning;

\noindent \textbf{(2)} We identify the \textbf{relative ranking of candidate trajectories} as a key attack target in world-model planning, moving beyond conventional objectives such as one-step predictions, latent representations, or policy outputs;

\noindent \textbf{(3)} We introduce a \textbf{tail-aware ranking objective} with dual gating mechanisms for stable and behavior-preserving manipulation of planning behavior.

\noindent \textbf{(4)} We validate TRAP on representative world models and control tasks, demonstrating strong attack effectiveness, clean-condition fidelity, and good transferability across planning paradigms.

\section{Related Work}

\subsection{World Models}
World models have become a central paradigm for model-based decision making by learning environment dynamics to support future prediction and planning~\cite{ha2018world,hafner2019learning,hafner2019dream,cheng2024scaling,li2024open,novelli2024operator,wu2025rlvr,zhang2025objects,luo2024survey}. Early work such as World Models~\cite{ha2018world} demonstrated that learned generative models could support policy learning, inspiring a broad line of research in model-based reinforcement learning. Among recent representative approaches, the Dreamer family~\cite{hafner2019dream,hafner2020mastering,hafner2023mastering}, culminating in DreamerV3~\cite{hafner2023mastering}, performs long-horizon policy optimization through latent imagination and has shown strong performance across diverse visual control and robotics tasks. In contrast, TD-MPC2~\cite{hansen2022temporal,hansen2023td} represents a different planning paradigm: instead of extended imagination-based policy optimization, it adopts a latent model predictive control framework for short-horizon rollout optimization and action selection. These two models therefore capture two representative forms of world-model-driven planning. Other efforts, such as SOLD~\cite{mosbach2024sold} and STORM~\cite{zhang2023storm}, further improve modeling capacity through object-centric dynamics representations and Transformer-based world models, respectively, while a parallel line of work studies video-generative world models~\cite{bruce2024genie,parker2024genie,brooks2024video,wang2024worlddreamer,qin2024worldsimbench,he2026pre,kang2024far}. In contrast to prior work that primarily aims to improve modeling and planning capability, our work investigates the security vulnerability of planning-based world models under backdoor attacks.

\subsection{Backdoor Attacks}
Backdoor attacks have been widely studied in visual classification models, where triggers are implanted during training to hijack predictions while preserving clean-data performance~\cite{gu2019badnets,li2022backdoor}. Similar vulnerabilities have since been extended to multimodal large language models and vision-language-action models as visual encoders become standard building blocks of foundation models~\cite{zhou2025badvla,xu2025tabvla,guo2026state,zhou2026inject}. In particular, BadVLA~\cite{zhou2025badvla} first exposed backdoor vulnerabilities in VLA policies, while follow-up works such as TabVLA~\cite{xu2025tabvla} and State Backdoor~\cite{guo2026state} explored more stealthy attack paradigms based on targeted behavior manipulation and state-space triggers. Related attacks have also been studied in reinforcement learning, primarily through training-time poisoning or trigger-conditioned manipulation of policy-based agents~\cite{kiourti2020trojdrl,cui2024badrl,gong2024baffle,ma2025unidoor}. In contrast, backdoor vulnerabilities in world models remain underexplored, especially those associated with long-horizon planning mechanisms. This gap directly motivates our work.

\subsection{Security of Model-Based Reinforcement Learning and World Models}

While reinforcement learning security has attracted growing attention, most existing work focuses on model-free algorithms, leaving the security of model-based RL relatively underexplored~\cite{lin2017tactics,pinto2017robust,kiourti2020trojdrl,cui2024badrl,bai2025rat}. A small body of recent work has begun to examine security risks in world models, primarily in the training phase. For example, data poisoning attacks manipulate interaction data to distort learned dynamics, causing imagined trajectories to deviate from true environmental transitions and inducing systematic planning biases~\cite{hustealthy}. However, such methods rely on compromising the training pipeline rather than enabling trigger-conditioned manipulation at deployment time. In contrast, our work studies an inference-time, trigger-conditioned attack setting that directly targets the planning mechanism of world models.

\section{Preliminaries}

\subsection{World-Model Planning}

We consider a world-model-driven agent that uses a learned dynamics model to internally simulate future outcomes and guide decision-making~\cite{ha2018world,hafner2019dream,hafner2023mastering,hansen2022temporal,hansen2023td}. Given the current observation $o_t$, the agent rolls out a set of candidate trajectories under the learned model and evaluates them using predicted rewards, values, or trajectory-level scores~\cite{hafner2019dream,hafner2023mastering,hansen2022temporal,hansen2023td,chen2022transdreamer,zhang2023storm}. The resulting action is then selected according to these imagined evaluations.

At time $t$, this process can be abstracted as selecting the action sequence that maximizes the planner's trajectory-level objective:
\[
a_{t:t+H-1}^{*}
=
\arg\max_{a_{t:t+H-1}}
J\bigl(o_t, a_{t:t+H-1}\bigr),
\]
where $H$ is the planning horizon and $J(\cdot)$ denotes the world-model-based evaluation of a candidate trajectory.

This formulation covers representative planning-based world models such as DreamerV3~\cite{hafner2023mastering} and TD-MPC2~\cite{hansen2023td}. Although these methods differ algorithmically---e.g., latent imagination-based policy optimization versus model predictive control~\cite{hafner2019dream,hafner2023mastering,hansen2022temporal,hansen2023td}---they share the same decision principle: behavior is determined by the comparative evaluation of model-generated future trajectories. This trajectory-level dependence forms the primary attack surface considered in our work.

\subsection{Threat Model}

\noindent\textbf{Attacker's Goal.}
The attacker aims to perform a trigger-conditioned, deployment-time backdoor attack. By attaching a small visual trigger to the observation, the attacker seeks to systematically distort the agent's long-horizon planning outcome and induce persistent behavioral deviation.

\noindent\textbf{Attacker's Knowledge.}
We consider a white-box setting in which the attacker has access to the target world model and its associated policy or planner. This assumption is realistic for open-source model-based RL systems and allows the attacker to analyze the internal trajectory-evaluation mechanism. In this work, we instantiate the threat model on DreamerV3~\cite{hafner2023mastering} and TD-MPC2~\cite{hansen2023td} as two representative examples, while the proposed framework is broadly applicable to any world-model-driven planner that relies on imagined trajectory evaluation.

\noindent\textbf{Attacker's Capability.}
The attacker cannot modify model parameters, alter the environment dynamics, or poison the original training pipeline. The attack is restricted to a spatially localized visual patch applied at deployment time~\cite{brown2017adversarial,karmon2018lavan,athalye2018synthesizing}. To limit perturbation magnitude, the trigger is constrained by an $\ell_{\infty}$ budget:
\[
\|\delta\|_{\infty} \leq \epsilon.
\]
The attacker seeks a universal trigger that reliably degrades decision quality whenever present. Importantly, our method does not require access to the original large-scale training dataset; only a small set of proxy trajectories collected from interaction with the target agent is needed for trigger optimization.

\subsection{Generic Patch-Based Attack Formulation}

Let $T(o_t,\delta)$ denote the operation of applying a trigger patch $\delta$ to observation $o_t$. The triggered observation is defined as
\[
\tilde{o}_t = T(o_t, \delta) = (1-M)\odot o_t + M\odot \delta,
\]
where $M$ is a binary mask specifying the patch location and shape~\cite{brown2017adversarial,karmon2018lavan,athalye2018synthesizing}. To satisfy the perturbation budget during optimization, we parameterize the patch as
\[
\delta = \epsilon \cdot \tanh(\rho),
\]
where $\rho$ is the unconstrained optimization variable.

The attacker therefore seeks a trigger that changes the planner's trajectory evaluation under $\tilde{o}_t$ while leaving clean behavior under $o_t$ largely unaffected. A natural objective is to reduce the model-predicted return under triggered inputs. Let $\hat{z}_{t+k}$ denote the imagined latent state at step $k$ under the triggered observation, and let the world model evaluate this rollout using predicted rewards and values, where $\gamma \in (0,1]$ is the discount factor and $\alpha$ weights the value estimate relative to the reward term. A generic deployment-time attack can then be written as
\[
\min_{\rho}\ \mathcal{L}_{\mathrm{atk}}
=
\mathbb{E}_{o_t \sim \mathcal{D}_{\mathrm{proxy}}}
\left[
\sum_{k=1}^{K}\gamma^k
\bigl(r(\hat{z}_{t+k}) + \alpha V(\hat{z}_{t+k})\bigr)
\right].
\]

However, as we show next, directly minimizing predicted return is
insufficient for reliably attacking world-model planning, because action
selection ultimately depends on the \emph{relative ranking} of candidate
trajectories rather than their absolute scores.

\begin{figure*}[t]
    \centering
    \includegraphics[width=0.96\textwidth, trim=1.2cm 4.5cm 1.2cm 2.0cm, clip]{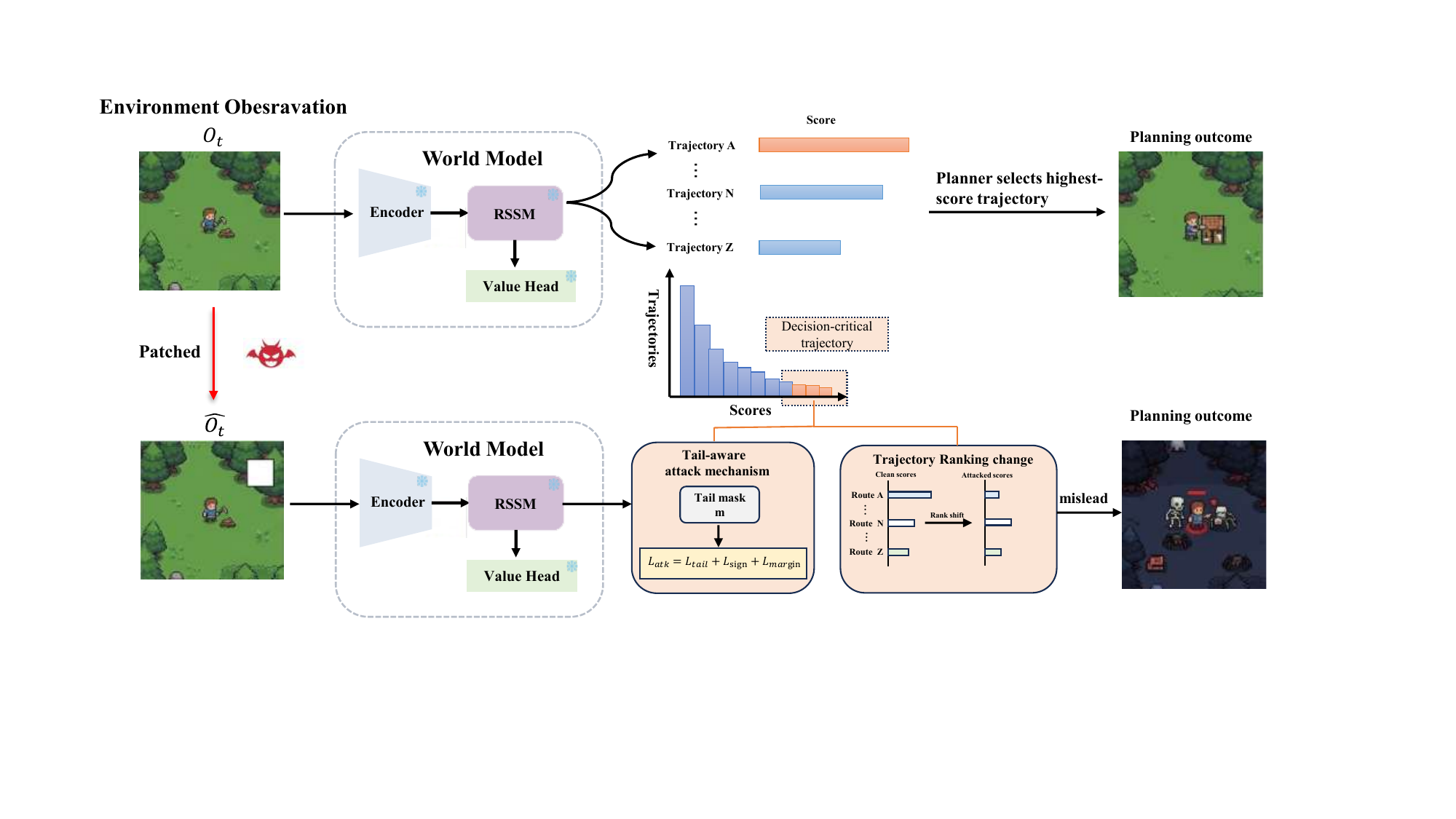}
    \caption{Overview of TRAP. The trigger selectively suppresses decision-critical tail trajectories, shifts trajectory ranking, and misleads world-model planning.}
    \label{fig:method_overview}
\end{figure*}

\section{Method}

This section presents \textbf{TRAP} (\textbf{T}ail-aware \textbf{R}anking \textbf{A}ttack on \textbf{P}lanning), our proposed inference-time backdoor framework against world models. TRAP targets a central decision mechanism of world-model-driven agents: the \emph{relative ranking} of candidate imagined trajectories. Building on the vulnerability analysis in Sec.~3, our key insight is that action selection is not determined by the average predictive quality across all imagined trajectories, but rather by the relative ordering of a small number of high-scoring ``tail'' trajectories. Therefore, an effective backdoor attack should not perturb all trajectories uniformly; instead, it should prioritize and consistently manipulate those high-value tail trajectories that are most critical to decision-making.

To instantiate this insight, the TRAP framework consists of two core components. The first is a \emph{tail-aware ranking loss} that operates directly on the high-value tail of imagined trajectories, systematically suppressing the relative scores of decision-critical trajectories. The second is a \emph{gated optimization mechanism}, implemented via auxiliary bounding losses, that filters out unstable or misaligned gradient updates, thereby promoting both attack stability and behavior preservation. We detail these components below.

\subsection{Identifying Tail Trajectories as the Decision Pivot}

Given the current observation $o_t$, a world-model-driven agent generates a set of imagined trajectories in latent space and evaluates them to guide action selection. Let $J_{b,k}(o_t)$ denote the evaluation score of the $b$-th imagined trajectory at imagination step $k$, where $b \in \{1, \dots, B\}$ indexes candidate trajectories and $k \in \{1, \dots, K\}$ denotes the planning horizon. Here, $J_{b,k}$ serves as a unified notation for the trajectory-level objective used by the planner, such as cumulative reward or expected state value.

Recall the patch injection operator defined in Sec.~3.3, which yields the triggered observation $\tilde{o}_t = T(o_t, \delta)$. Under clean and triggered conditions, the world model produces two corresponding sets of trajectory scores:
\[
J_{b,k}^{\mathrm{clean}} = J_{b,k}(o_t),
\qquad
J_{b,k}^{\mathrm{trig}} = J_{b,k}(\tilde{o}_t).
\]
We define the trigger-induced score deviation as
\[
\Delta J_{b,k} = J_{b,k}^{\mathrm{trig}} - J_{b,k}^{\mathrm{clean}}.
\]

The attacker's goal is to optimize the trigger $\delta$ such that high-scoring trajectories, which are most likely to dominate planning, are systematically pushed down in the ranking. Therefore, rather than optimizing a uniform objective over the full set of imagined trajectories, TRAP isolates the subset of trajectories that lie in the high-score tail of the clean distribution. This distinction is crucial: because these top-tier tail trajectories contribute disproportionately to final action selection, targeted ranking shifts within this subset can disproportionately distort the agent's ultimate decision.

\subsection{Tail-Aware Ranking Loss}

Building on the analysis in Sec.~4.1, imagined trajectory scores in world models often exhibit a pronounced long-tailed distribution: most trajectories lie in the low-score region, while only a small high-score tail dominates the final decision. We therefore construct a tail mask based on the clean trajectory scores and restrict the primary attack objective to these decision-critical trajectories.

Specifically, for each trajectory $b \in \{1, \dots, B\}$, we identify the tail index set along the planning horizon:
\[
\mathcal{T}_b = \mathrm{Tail}\!\left(\{J_{b,k}^{\mathrm{clean}}\}_{k=1}^{K}\right),
\]
where $\mathrm{Tail}(\cdot)$ denotes a top-quantile selection operator that retains the highest-scoring imagined steps under the clean condition. We then define the corresponding binary mask
\[
M_{b,k} =
\begin{cases}
1, & k \in \mathcal{T}_b,\\
0, & \text{otherwise}.
\end{cases}
\]

Using this mask, we define the tail attack score for each trajectory as
\[
S_b = \rho_K \!\left( \{\Delta J_{b,k}\}_{k=1}^{K} \right)
=
\frac{\sum_{k=1}^{K} M_{b,k}\,\Delta J_{b,k}}
{\sum_{k=1}^{K} M_{b,k} + \varepsilon},
\]
where $\varepsilon$ is a numerical stabilizer, and $\rho_K(\cdot)$ denotes the inner aggregation operator over the masked horizon. Since $\Delta J_{b,k} = J_{b,k}^{\mathrm{trig}} - J_{b,k}^{\mathrm{clean}}$, a negative value of $S_b$ indicates that the trigger successfully suppresses the scores of the critical high-value steps in trajectory $b$, whereas $S_b > 0$ implies that the attack is ineffective on this trajectory.

To form the final tail-aware loss, we further apply an outer aggregation operator $\rho_B(\cdot)$ across candidate trajectories:
\[
\mathcal{L}_{\mathrm{tail}}
=
\rho_B\!\Bigl(
\bigl\{
\rho_K(\{\Delta J_{b,k}\}_{k=1}^{K})
\bigr\}_{b=1}^{B}
\Bigr).
\]

In our implementation, we instantiate $\rho_B$ as a softmin operator over the per-trajectory tail attack scores $\{S_b\}_{b=1}^{B}$. Compared with a naive mean aggregation, softmin places greater emphasis on trajectories whose critical tail scores remain relatively difficult to suppress, while retaining smoother gradients than a hard minimum. Therefore, the tail-aware component of TRAP is defined by selecting the clean-condition high-score tail within each trajectory and then aggregating the resulting per-trajectory attack scores with a softmin reducer across trajectories.

From an optimization perspective, minimizing $\mathcal{L}_{\mathrm{tail}}$ selectively suppresses the subset of trajectories most likely to dominate clean planning, thereby systematically shifting the relative ranking of imagined trajectories. This constitutes the core mechanism of TRAP.

\subsection{Gated Optimization for Stable Ranking Manipulation}

Optimizing the tail-aware loss alone introduces an important challenge. Because imagined planning is high-variance and strongly coupled across rollout steps, directly suppressing tail scores can easily destabilize optimization. In practice, this leads to two failure modes. First, some trajectory deviations $\Delta J_{b,k}$ may become positive, indicating that the trigger is moving in the wrong direction and inadvertently increasing scores that should be suppressed. Second, some deviations may already be sufficiently negative, in which case further suppression is unnecessary and may make optimization overly aggressive.

To address these issues, we introduce a \emph{dual gating mechanism} that stabilizes optimization from two complementary perspectives: \emph{directional control} and \emph{magnitude control}.

\subsubsection{Sign Gate: Directional Control}

We first introduce a \emph{sign gate} to penalize trajectory entries whose scores are not successfully suppressed and may instead be increased. Recall the score deviation
\[
\Delta J_{b,k} = J_{b,k}^{\mathrm{trig}} - J_{b,k}^{\mathrm{clean}}.
\]
Our objective requires $\Delta J_{b,k} < 0$, so that the trigger reduces the score of decision-critical imagined trajectories. When $\Delta J_{b,k} > 0$, the optimization has deviated from the intended attack direction. We therefore define the sign gate as
\[
G_{b,k}^{\mathrm{sign}} = \mathbf{1}(\Delta J_{b,k} > 0),
\]
and construct the corresponding sign-gated loss:
\[
\mathcal{L}_{\mathrm{sign}} =
\frac{1}{BK}
\sum_{b=1}^{B}\sum_{k=1}^{K}
G_{b,k}^{\mathrm{sign}}\,(\Delta J_{b,k})^2.
\]
This loss is activated only when the update moves in the wrong direction, and therefore does not interfere with trajectory scores that have already been successfully reduced.

\subsubsection{Magnitude Gate: Amplitude Control}

Directional control alone is insufficient. If optimization is overly aggressive, tail scores may be pushed down more than necessary, leading to unnecessarily large score distortion and reduced optimization stability.

To avoid this, we further introduce a \emph{magnitude gate} to restrict excessively large negative deviations. Specifically, we define a safety margin $\beta$ and regard a trajectory entry as already sufficiently suppressed when
\[
\Delta J_{b,k} < -\beta.
\]
In this case, further suppression is unnecessary. The corresponding gate is defined as
\[
G_{b,k}^{\mathrm{mag}} = \mathbf{1}(\Delta J_{b,k} < -\beta),
\]
and the resulting magnitude-constrained loss is
\[
\mathcal{L}_{\mathrm{mag}} =
\frac{1}{BK}
\sum_{b=1}^{B}\sum_{k=1}^{K}
G_{b,k}^{\mathrm{mag}}\,(\Delta J_{b,k} + \beta)^2.
\]
This term penalizes overly large negative deviations and prevents unnecessarily aggressive updates, thereby keeping optimization in a more stable regime.

\begin{table*}[!t]
\caption{Main results of the random patch baseline and \textsc{TRAP} on two representative world-model paradigms, DreamerV3 and TD-MPC2. We report Mean Return Drop and Attack Success Rate (ASR), where higher values indicate stronger attack effectiveness. Unless otherwise noted, all results are obtained with patch ratio $0.09$ and perturbation budget $\epsilon=64$.}
\label{tab:main_baseline_vs_trap}
\centering
\small
\renewcommand{\arraystretch}{1.08}
\setlength{\tabcolsep}{5pt}
\begin{tabular}{lllcccc}
\toprule
\multirow{2}{*}{Model} & \multirow{2}{*}{Env.} & \multirow{2}{*}{Task} 
& \multicolumn{2}{c}{Baseline} & \multicolumn{2}{c}{\textsc{TRAP}} \\
\cmidrule(lr){4-5} \cmidrule(lr){6-7}
& & & Mean Drop (\%) $\uparrow$ & ASR (\%) $\uparrow$ 
    & Mean Drop (\%) $\uparrow$ & ASR (\%) $\uparrow$ \\
\midrule

\multirow{9}{*}{DreamerV3}
& Crafter    & crafter        & -0.1 $\pm$ 3.5  & 43.9 $\pm$ 8.4         & \textbf{63.2 $\pm$ 0.9}  & \textbf{98.1 $\pm$ 0.8} \\
& DMControl & cheetah-run    & -0.4 $\pm$ 4.2  & 45.8 $\pm$ 6.3         & \textbf{22.8 $\pm$ 2.3}  & \textbf{99.6 $\pm$ 0.5} \\
& DMControl & dog-run        & -0.3 $\pm$ 3.8  & 50.0 $\pm$ 6.1           & \textbf{9.8 $\pm$ 3.8}  & \textbf{77.2 $\pm$ 3.7} \\
& DMControl & humanoid-walk  & 3.1 $\pm$ 6.4  & 65.3 $\pm$ 5.5       & \textbf{69.8 $\pm$ 0.7}  & \textbf{100.0 $\pm$ 0.0} \\
& DMControl & walker-walk    & -0.2 $\pm$ 5.8  & 48.0 $\pm$ 7.2         & \textbf{18.5 $\pm$ 1.5}  & \textbf{100.0 $\pm$ 0.0} \\
& Atari      & seaquest       & 97.2 $\pm$ 1.9 & \textbf{100.0 $\pm$ 0.0} & \textbf{97.5 $\pm$ 0.8}  & \textbf{100.0 $\pm$ 0.0} \\
& Atari      & pong           & 151.8 $\pm$ 11.6 & \textbf{100.0 $\pm$ 0.0} & \textbf{164.5 $\pm$ 8.3} & \textbf{100.0 $\pm$ 0.0} \\
& Atari      & breakout       & 97.1 $\pm$ 1.4 & \textbf{100.0 $\pm$ 0.0} & \textbf{98.8 $\pm$ 0.7}  & \textbf{100.0 $\pm$ 0.0} \\
& Atari      & invaders       & 90.4 $\pm$ 3.0 & \textbf{100.0 $\pm$ 0.0} & \textbf{93.3 $\pm$ 1.8}  & \textbf{100.0 $\pm$ 0.0} \\
\midrule

\multirow{3}{*}{TD-MPC2}
& DMControl & hopper-hop     & 51.2 $\pm$ 4.0 & 95.0 $\pm$ 1.7          & \textbf{99.8 $\pm$ 0.3} & \textbf{100.0 $\pm$ 0.0} \\
& DMControl & cheetah-run    & 9.3 $\pm$ 5.3  & 90.0 $\pm$ 2.3         & \textbf{98.4 $\pm$ 0.4}  & \textbf{100.0 $\pm$ 0.0} \\
& DMControl & walker-walk    & 1.9 $\pm$ 3.2   & 64.0 $\pm$ 2.5          & \textbf{92.4 $\pm$ 0.9}  & \textbf{100.0 $\pm$ 0.0} \\
\bottomrule
\end{tabular}
\end{table*}

\begin{table*}[!t]
\caption{Detailed clean and patched episodic returns, together with evaluation efficiency statistics, for DreamerV3 and TD-MPC2 across different tasks. Tasks are grouped by environment category. Eval Time is reported in seconds for 100 evaluation episodes, and runtime overhead is measured relative to clean evaluation. The corresponding relative degradation metrics (Mean Return Drop and ASR) are summarized in Table~\ref{tab:main_baseline_vs_trap}.}
\label{tab:appendix_dreamer_tdmpc2}
\centering
\small
\renewcommand{\arraystretch}{1.08}
\setlength{\tabcolsep}{6pt}
\begin{tabular}{lllcccc}
\toprule
Model & Environment & Task & Clean Return (mean$\pm$std) $\uparrow$ & Patched Return (mean$\pm$std) $\downarrow$ & Eval Time $\downarrow$ & Runtime Overhead $\downarrow$ \\
\midrule

\multirow{9}{*}{DreamerV3}
& \multirow{4}{*}{DMControl}
& cheetah-run    & 889.44 $\pm$ 23.91    & 656.14 $\pm$ 186.41   & 1808.1  & 1.010$\times$ \\
& & dog-run        & 368.63 $\pm$ 40.53    & 330.15 $\pm$ 32.36    & 3131.9  & 1.006$\times$ \\
& & humanoid-walk  & 800.00 $\pm$ 34.73    & 231.77 $\pm$ 57.55    & 2216.2  & 1.009$\times$ \\
& & walker-walk    & 966.63 $\pm$ 15.93    & 787.79 $\pm$ 49.17    & 1683.2  & 1.015$\times$ \\
\cmidrule(lr){2-7}
& Crafter
& crafter        & 12.91 $\pm$ 1.88      & 4.61 $\pm$ 2.22       & 588.6   & 0.812$\times$ \\
\cmidrule(lr){2-7}
& \multirow{4}{*}{Atari}
& pong           & 20.31 $\pm$ 0.88      & -13.09 $\pm$ 3.44     & 2225.2  & 0.960$\times$ \\
& & breakout       & 293.50 $\pm$ 153.64   & 1.95 $\pm$ 1.63       & 2535.7  & 0.166$\times$ \\
& & seaquest       & 6611.20 $\pm$ 1255.86 & 165.20 $\pm$ 79.73    & 11386.2 & 0.177$\times$ \\
& & space-invaders & 8787.85 $\pm$ 6591.83 & 586.10 $\pm$ 136.06   & 2765.9  & 0.159$\times$ \\
\midrule

\multirow{3}{*}{TD-MPC2}
& \multirow{3}{*}{DMControl}
& cheetah-run    & 635.28 $\pm$ 27.16    & 10.38 $\pm$ 4.29      & 3690.5  & 1.202$\times$ \\
& & hopper-hop     & 304.23 $\pm$ 12.94    & 0.00 $\pm$ 0.02       & 3663.4  & 1.270$\times$ \\
& & walker-walk    & 951.23 $\pm$ 25.44    & 72.71 $\pm$ 23.38     & 3845.0  & 1.214$\times$ \\
\bottomrule
\end{tabular}
\end{table*}

\subsection{Overall Attack Objective and Optimization Procedure}

Combining the tail-aware ranking loss, the dual gating mechanism, and the image-space regularizers, our final TRAP objective is defined as
\[
\mathcal{L}_{\mathrm{atk}}
=
\mathcal{L}_{\mathrm{tail}}
+
\mathcal{L}_{\mathrm{sign}}
+
\mathcal{L}_{\mathrm{mag}}
+
\mathcal{L}_{\mathrm{tv}}
+
\mathcal{L}_{\mathrm{l2}}.
\]
For clarity, we omit the scalar weighting coefficients; in practice, the auxiliary terms use fixed coefficients across experiments. Here, $\mathcal{L}_{\mathrm{tail}}$ drives the core ranking manipulation objective, $\mathcal{L}_{\mathrm{sign}}$ penalizes updates in the wrong direction, and $\mathcal{L}_{\mathrm{mag}}$ constrains overly aggressive score suppression. In addition, $\mathcal{L}_{\mathrm{tv}}$ and $\mathcal{L}_{\mathrm{l2}}$ act as image-space regularizers that encourage spatial smoothness and control the patch energy, respectively, thereby regularizing patch structure and improving optimization stability. In practice, we use
\[
\mathcal{L}_{\mathrm{tv}} = \mathrm{TV}(\delta),
\qquad
\mathcal{L}_{\mathrm{l2}} = \|\delta\|_2^2.
\]

Recall from Sec.~3.2 that attack optimization does not require online interaction with the environment during optimization. Instead, the attacker relies on a small set of proxy trajectories, denoted by $\mathcal{D}_{\mathrm{proxy}}$, which can be collected by deploying the pretrained model in the environment for a few episodes. This design is consistent with the deployment-time threat setting considered in this work.

During optimization, rather than relying on projected gradient descent with explicit clipping, we use the smooth reparameterization introduced in Sec.~3.3:
\[
\delta = \epsilon \cdot \tanh(\rho).
\]
We then optimize the unconstrained variable $\rho$ using gradient-based updates:
\[
\rho \leftarrow \rho - \eta \nabla_{\rho}\mathcal{L}_{\mathrm{atk}},
\]
where $\eta$ is the learning rate. At each optimization step, we perform imagined rollouts for a batch of proxy observations under both clean and triggered conditions, construct the tail mask $\mathcal{T}_b$ according to the clean trajectory scores, evaluate the overall objective $\mathcal{L}_{\mathrm{atk}}$, and update $\rho$ accordingly. By repeating this process, the patch gradually learns to shift the relative ranking of decision-critical imagined trajectories.

TRAP does not maximize raw prediction error, but instead targets the ranking structure induced by high-value tail trajectories. This explains why it can reliably disrupt long-horizon planning even when one-step prediction errors remain small.

\section{Experiments}

\subsection{Experimental Setup}

\noindent\textbf{Evaluated Models.}
We evaluate the proposed attack on two representative classes of world-model frameworks. The first is \emph{latent imagination-based world models}, represented by DreamerV3~\cite{hafner2023mastering}, which perform policy learning through latent dynamics modeling and imagined rollouts. The second is \emph{planning-based world-model control frameworks}, represented by TD-MPC2~\cite{hansen2023td}, which combine learned world models with online planning and select actions through trajectory evaluation. These two systems represent two major paradigms of world-model decision-making: imagination-driven policy learning and model-based planning control. Evaluating both helps demonstrate that TRAP is not tied to a specific implementation, but instead targets a ranking-driven decision mechanism shared across world-model-based agents.

\noindent\textbf{Environments.}
We conduct experiments on a diverse set of reinforcement learning environments covering dense-reward control, sparse-reward planning, and arcade-style decision making. Specifically, we use several continuous-control tasks from the DeepMind Control Suite~\cite{tassa2018deepmind} to evaluate attack effectiveness in high-frequency control settings, Crafter~\cite{hafner2021benchmarking} to assess robustness in sparse-reward and long-horizon planning scenarios, and Atari games~\cite{bellemare2013arcade} to test whether the attack remains effective in visually rich discrete-action environments. This diversity allows us to examine TRAP across substantially different planning horizons, reward structures, and trajectory distributions.

\noindent\textbf{Attack Setting.}
We adopt a patch-based backdoor setting in which a localized, fixed, and budget-constrained trigger is overlaid onto the input observation, and abnormal behavior is activated only when the trigger is present~\cite{brown2017adversarial,karmon2018lavan,athalye2018synthesizing,gu2019badnets}. To regularize the spatial structure of the learned patch, the patch is constrained in both spatial extent and perturbation magnitude, and we further incorporate Total Variation (TV) regularization during optimization, with $\lambda_{\mathrm{tv}} = 10^{-3}$. During attack optimization, only the patch parameters are updated, while the target world model remains unchanged. Unless otherwise specified, the default attack setting uses patch ratio $0.09$ and perturbation budget $\epsilon=64$. During evaluation, we compare agent behavior under both clean and triggered conditions. For each task, we repeat trigger optimization across multiple independent runs and report mean and standard deviation to assess attack robustness.

\begin{figure*}[t]
    \centering
    \includegraphics[width=\textwidth]{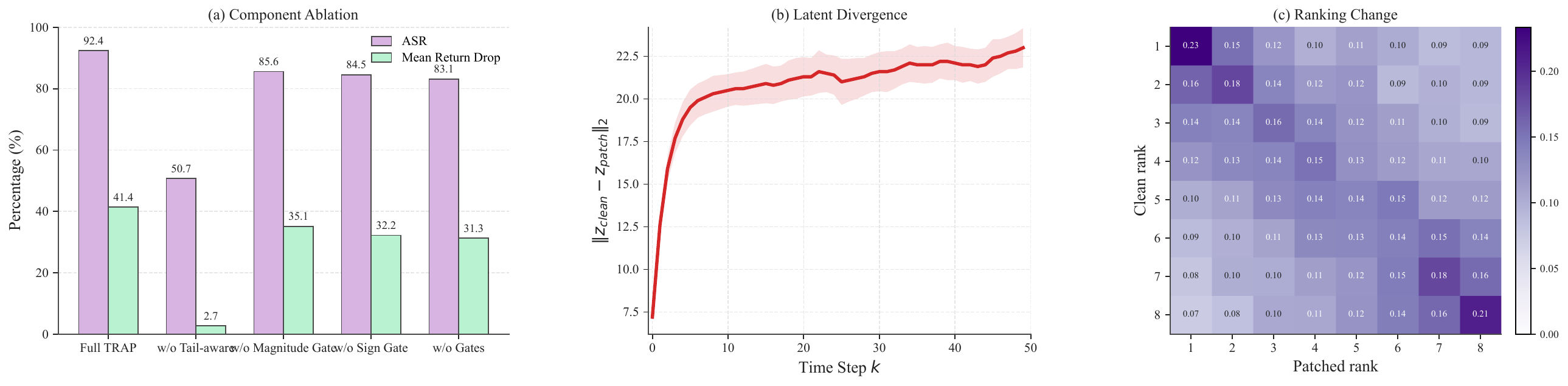}
    \caption{Ablation and mechanistic analysis of TRAP. (a) Component ablation under $\epsilon=32$, reporting Attack Success Rate (ASR) and Mean Return Drop. (b) Latent divergence between clean and triggered imagination rollouts, measured as $\|z_k^{\mathrm{clean}}-z_k^{\mathrm{trig}}\|_2$ over imagination step $k$. (c) Clean-to-triggered rank transition of imagined trajectories. Diagonal concentration indicates preserved ordering, while off-diagonal mass reflects trigger-induced rank shifts in the planner's trajectory ranking.}
    \label{fig:ablation_mech}
\end{figure*}

\noindent\textbf{Metrics.}
We use \emph{Mean Return Drop} as the primary effectiveness metric, defined as the relative percentage decrease in episodic return under the trigger compared with the clean condition. This metric captures the overall extent to which the attack degrades actual decision quality in the environment. We also report \emph{Attack Success Rate (ASR)}, defined as the fraction of episodes in which the triggered return is lower than the corresponding clean return, in order to measure attack consistency across episodes. Compared with average return drop alone, ASR provides a clearer view of whether degradation is persistent rather than driven by a small number of extreme failures. In addition, we report evaluation-time overhead to quantify the extra computational cost introduced by the attack.

\subsection{Main Results}

Table~\ref{tab:main_baseline_vs_trap} summarizes the main comparison between \textsc{TRAP} and a random patch baseline on DreamerV3 and TD-MPC2 across multiple tasks. Overall, \textsc{TRAP} consistently outperforms the random baseline on both world-model paradigms, yielding higher Mean Return Drop and higher Attack Success Rate (ASR) in nearly all settings. This suggests that the gains of \textsc{TRAP} are not explained by patch injection alone, and are consistent with its targeted manipulation of ranking-driven planning in world-model-based agents.

On DreamerV3, \textsc{TRAP} leads to substantially stronger degradation than the random baseline on Crafter and most DMControl tasks. In particular, on Crafter, the random patch causes almost no average degradation ($-0.1 \pm 3.5\%$ Mean Return Drop), whereas \textsc{TRAP} increases this value to $63.2 \pm 0.9\%$ and raises ASR from $43.9 \pm 8.4\%$ to $98.1 \pm 0.8\%$. Similarly, on DMControl tasks such as \textit{humanoid-walk} and \textit{walker-walk}, \textsc{TRAP} produces much larger return drops while also driving ASR to $100.0 \pm 0.0\%$. Even on Atari tasks where the random patch baseline is already highly disruptive, \textsc{TRAP} remains competitive, although its margin over the baseline is small in these near-ceiling settings. This suggests that some Atari environments may already be broadly sensitive to patch perturbations, regardless of how the trigger is optimized.

The gains are even more pronounced on TD-MPC2. Across all three evaluated DMControl tasks, \textsc{TRAP} yields large return drops together with near-perfect or perfect ASR. For example, on \textit{cheetah-run}, Mean Return Drop rises from $9.3 \pm 5.3\%$ under the random patch baseline to $98.4 \pm 0.4\%$ under \textsc{TRAP}, and on \textit{walker-walk} it rises from $1.9 \pm 3.2\%$ to $92.4 \pm 0.9\%$. Together, these results indicate that \textsc{TRAP} transfers effectively across both latent imagination-based and planning-based world-model paradigms.

Table~\ref{tab:appendix_dreamer_tdmpc2} further reports absolute returns and runtime statistics, confirming that the degradation in Table~\ref{tab:main_baseline_vs_trap} reflects genuine degradation in long-horizon decision quality rather than metric artifacts, while incurring only modest computational overhead.

\begin{figure*}[t]
    \centering
    \includegraphics[width=\textwidth]{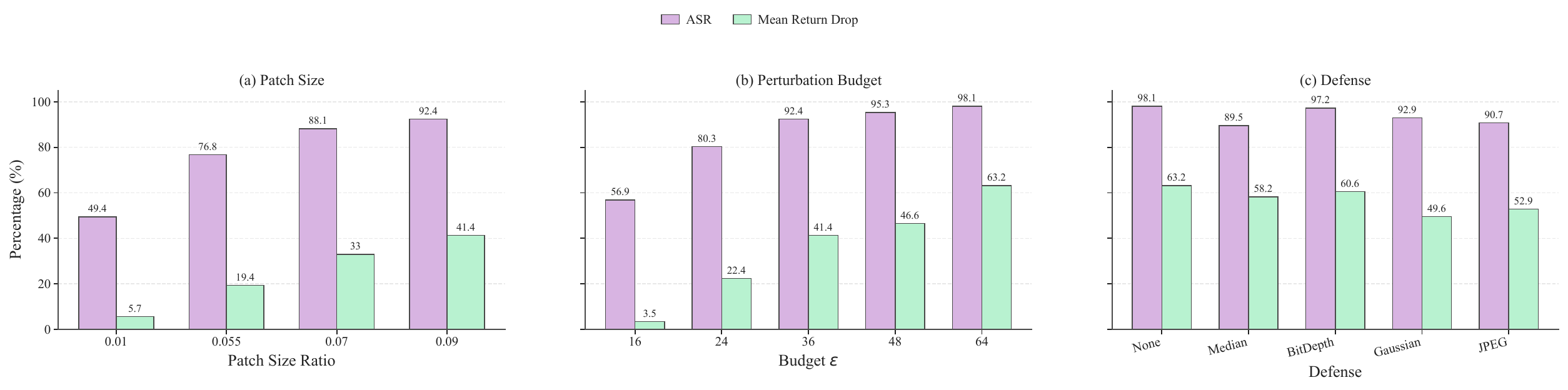}
    \caption{\textbf{Diagnostic experiments and defense analysis of TRAP.}
(a) Effect of patch size ratio on Attack Success Rate (ASR) and Mean Return Drop.
(b) Effect of perturbation budget $\epsilon$ on attack effectiveness.
(c) Attack effectiveness under several standard image-space defenses.}
    \label{fig:diag_patch_defense}
\end{figure*}

\subsection{Ablation and Mechanistic Analysis}

We next analyze which components of TRAP are responsible for its effectiveness and why the attack remains effective over long-horizon imagination. Fig.~\ref{fig:ablation_mech}(a) reports the component ablation results under a moderately constrained setting ($\epsilon=32$). We use this budget to avoid ceiling effects at larger perturbation levels, so that the contributions of individual components remain more clearly distinguishable. Under this setting, the full TRAP objective achieves the strongest overall performance, reaching $92.4\%$ ASR and $41.4\%$ Mean Return Drop.

The two gating components provide an additional but complementary benefit. Removing the magnitude gate reduces performance to $85.6\%$ ASR and $35.1\%$ Mean Return Drop, while removing the sign gate yields $84.5\%$ ASR and $32.2\%$ Mean Return Drop. Removing both gates further reduces performance to $83.1\%$ ASR and $31.3\%$ Mean Return Drop. Notably, the attack remains partially effective even without the gates, whereas removing the tail-aware objective almost collapses the attack. This indicates that the tail-aware ranking loss is the primary driver of TRAP, while the dual-gate design mainly improves optimization stability and strengthens the final attack effect by suppressing ineffective or overly aggressive updates.

To further understand the attack mechanism, Fig.~\ref{fig:ablation_mech}(b) tracks the latent divergence between clean and triggered imagination rollouts, measured as $\|z_k^{\mathrm{clean}} - z_k^{\mathrm{trig}}\|_2$ over imagination step $k$. The divergence increases rapidly in the first few rollout steps and then remains consistently large throughout the planning horizon. This trend indicates that the trigger does not merely induce a transient one-step perturbation; instead, it injects a persistent shift into the latent rollout process, which is subsequently propagated by the learned dynamics. The sustained separation between clean and triggered latent trajectories explains why the attack effect is not washed out during imagination, but can continue influencing downstream trajectory evaluation over long horizons.

Fig.~\ref{fig:ablation_mech}(c) provides more direct evidence of the decision-level effect of TRAP by visualizing the clean-to-triggered rank transition of imagined trajectories. If the trigger only caused uniform score degradation, the ranking would remain largely unchanged and the transition mass would stay concentrated on the diagonal. Instead, we observe substantial off-diagonal transitions, indicating that the trigger systematically alters the relative ordering of imagined trajectories. This result supports our central claim that TRAP succeeds not merely by perturbing latent rollouts, but by disrupting the ranking structure on which world-model planning depends.

Taken together, these results provide a coherent explanation of TRAP's effectiveness. The tail-aware objective focuses the attack on the subset of trajectories that most strongly influence decision-making, the gating terms stabilize optimization, and the trigger-induced perturbation persists throughout imagination before translating into direct disruption of trajectory ordering. Together, these effects enable reliable manipulation of long-horizon planning outcomes.

\subsection{Diagnostic Experiments}

We further examine how TRAP behaves under different patch sizes and perturbation budgets. Fig.~\ref{fig:diag_patch_defense}(a) shows the effect of varying the patch size ratio. As the patch becomes larger, both Attack Success Rate (ASR) and Mean Return Drop increase steadily, indicating that a larger spatial footprint gives the trigger greater influence over imagined trajectory evaluation. Notably, even relatively small patches already induce non-trivial performance degradation, while the default setting ($0.09$) achieves the strongest overall effect among the tested values.

Fig.~\ref{fig:diag_patch_defense}(b) examines the perturbation budget $\epsilon$. We observe a similarly monotonic trend: increasing $\epsilon$ consistently improves both ASR and Mean Return Drop. The attack remains effective even at moderate budgets and becomes particularly strong at larger budgets. This behavior is consistent with the design of TRAP, as a larger perturbation budget allows the trigger to induce stronger and more persistent distortions in latent rollout evaluation.

We also observe moderate sensitivity to patch position, although the attack remains effective across all tested locations. Overall, these diagnostics indicate that TRAP remains effective across a broad range of spatial footprints and perturbation budgets, rather than depending on a narrow operating regime.

\subsection{Defense Analysis}

We finally evaluate TRAP under several standard image-space defenses~\cite{xu2017feature,guo2017countering,karmon2018lavan,gushchin2024guardians,xue2023compression}. As shown in Fig.~\ref{fig:diag_patch_defense}(c), while all tested defenses reduce attack effectiveness to varying degrees, none is able to eliminate the attack entirely. Median filtering and JPEG compression moderately lower both ASR and Mean Return Drop relative to the undefended setting, whereas Gaussian smoothing leads to a more pronounced reduction in Mean Return Drop. In contrast, bit-depth reduction has only a limited effect, with performance remaining close to the undefended case.

Overall, these results suggest that TRAP does not merely exploit fragile, high-frequency pixel artifacts that can be easily removed by simple preprocessing. Instead, its attack effect remains substantial even after common image transformations. This indicates that the learned trigger is sufficiently robust to survive lightweight defenses and continue influencing the ranking of imagined trajectories.

\section{Conclusion}

We presented TRAP, a trigger-conditioned, inference-time backdoor attack framework against world-model-driven agents. TRAP targets a central decision-making mechanism in world models: the relative ranking of imagined trajectories. Our key insight is that world-model planning depends not on the average quality of all imagined futures, but on the relative ordering of a small subset of high-scoring tail trajectories that dominate final action selection. Based on this observation, TRAP combines a tail-aware ranking objective with dual gating mechanisms to manipulate planning outcomes in a stable manner while preserving clean-condition behavior.

Extensive experiments on DreamerV3 and TD-MPC2 across diverse tasks show that TRAP consistently achieves strong attack effectiveness while preserving clean-condition fidelity and incurring only modest computational overhead. Our ablation and mechanistic analyses further indicate that tail-aware ranking manipulation is the key contributor to attack success, while the gating components improve optimization stability. Overall, these findings highlight the need for dedicated security evaluation, robustness analysis, and defense design for world-model-based agents before their broader deployment in safety-critical settings.

\nocite{*}
\bibliographystyle{ACM-Reference-Format}
\bibliography{ref}

\end{document}